\documentclass[letterpaper, 10 pt, conference]{ieeeconf}  

\IEEEoverridecommandlockouts                              

\usepackage{graphics} 
\usepackage{epsfig} 
\usepackage{mathptmx} 
\usepackage{times} 
\usepackage{amsmath} 
\usepackage{amssymb}  
\usepackage{comment}
\usepackage{tabularx}
\usepackage{cite}

\usepackage[draft,nomargin,inline]{fixme}
\fxsetup{theme=color}

\usepackage{url}
\makeatletter
\let\NAT@parse\undefined
\makeatother
\usepackage{hyperref}

\title{\LARGE \bf
Semantic Interaction in Augmented Reality Environments\\for Microsoft HoloLens
}

\author{Peer Sch\"utt, Max Schwarz and Sven Behnke
\thanks{All authors are with Autonomous Intelligent Systems Group, Computer Science VI, University of Bonn, Germany, {\tt s6peschu@uni-bonn.de, \{schwarz, behnke\}@ais.uni-bonn.de.} This work was funded by grant BE 2556/16-1 (Research Unit FOR 2535 Anticipating Human Behavior) of the German Research Foundation (DFG).
\newline 978-1-7281-3605-9/19/\$31.00 \textcopyright 2019 IEEE}%
}

\begin{document}

\maketitle

\thispagestyle{empty}
\pagestyle{empty}

\begin{abstract}

Augmented Reality is a promising technique for human-machine interaction.
Especially in robotics, which always considers systems in their environment,
it is highly beneficial to display visualizations and receive user input
directly in exactly that environment.
We explore this idea using the Microsoft HoloLens, with which we capture
indoor environments and display interaction cues with known object classes.
The 3D mesh recorded by the HoloLens is annotated on-line, as the user moves,
with semantic classes
using a projective approach, which allows us to use a state-of-the-art 2D semantic
segmentation method. The results are fused onto the mesh; prominent object
segments are identified and displayed in 3D to the user.
Finally, the user can trigger actions by gesturing at the object.
We both present qualitative results and analyze the accuracy and performance
of our method in detail on an indoor dataset.
\end{abstract}

\section{Introduction}

Head-Mounted Displays (HMDs) like the HTC Vive, the Oculus Rift, Immersive Headsets or the Microsoft HoloLens are a branch of technology whose application areas and usage figures have been steadily increasing for years. Advances in technology have created new applications for such Virtual Reality (VR) and Augmented Reality (AR) devices, and their use in science and industry as tools and research objects is increasing.
In a robotics context, these devices allow 3D visualization of a robot's
perception of its environment and its planning, while also allowing a human user
of the system to give input in the 3D environment.
For this reason, this paper will take a closer look at the Microsoft HoloLens in order to examine the feasibility of selected concepts and investigate their usability in future research projects.

While increasingly better cameras, localization techniques and processors are being used to detect user movements and the environment, allowing them to be used as input for applications,
the developed methods so far lack the semantic understanding of the environment
necessary to really understand user intent (``Pick up that water bottle'').
The lack of this functionality will be addressed in this paper to explore ways to integrate this important factor into the HoloLens. We note that current state-of-the-art applications often are only
capable of tracking, i.e. require manual placement of a holographic object by
the user or fixed spatial markers at which the holograms are displayed.

The use of segmentation techniques is one way to provide applications with dynamically adaptable semantics.
The goal of semantic segmentation is to assign the appropriate class to each pixel of the image. Since each pixel is assigned a class, it is called a dense prediction, which is important as a substantial pre-processing step for advanced applications such as scene parsing or scene understanding. This technique is already widely used, e.g. in robotics, autonomous vehicles and various vision applications. However, the fields of application are becoming more and more diverse as the recognition and integration of semantics becomes more and more essential and opens up new fields of application. We chose RefineNet \cite{c9} as the CNN for semantic segmentation. This method has delivered state-of-the-art results on the Cityscapes dataset and has already been used successfully several times
in robotic applications \cite{c13, c16}.

In this work, RGB images are to be semantically segmented in order to expand them into a 3D scene by means of a fusion. On the basis of this segmentation in 3D, the interaction with recognized objects will be enabled in connection with the augmented reality capabilities of the HoloLens. This method corresponds to the current state of art, since there the segmentation of 3D scenes usually takes place via the regression of the segmentation into 2D, followed by a backprojection with a fusion into 3D.

\begin{figure}[t]
\centering
\includegraphics[width = \linewidth]{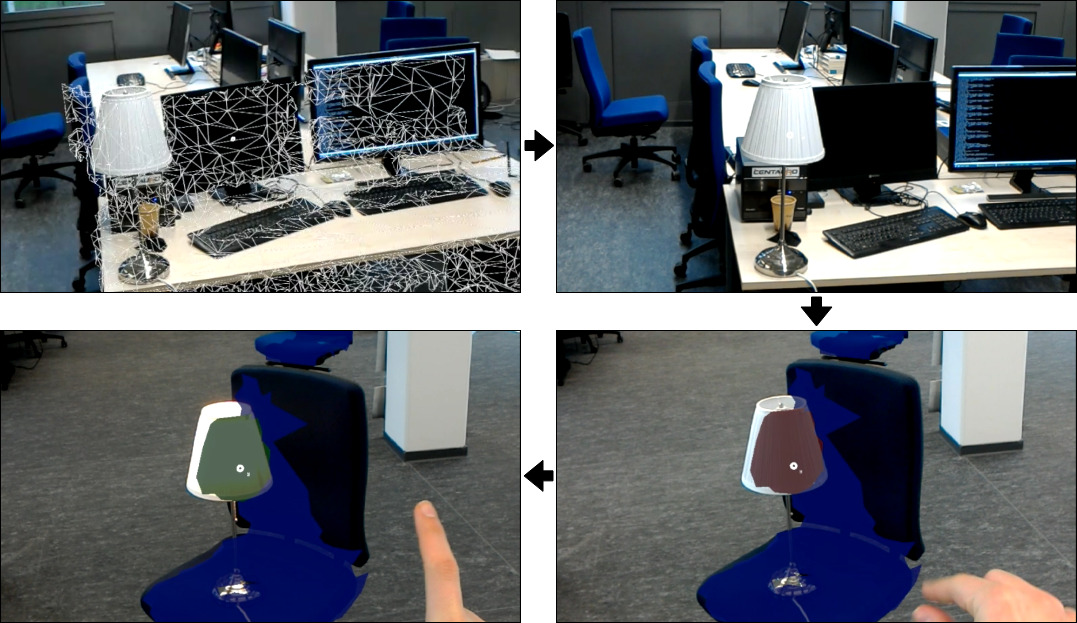}
\caption{Semantic Interaction. The HoloLens scans the surrounding (top left), takes images of it (top right) and presents segmentation results (bottom right). The user switches on a table-lamp (bottom left).}
\label{fig:4Frames_start}
\end{figure}

Our contributions presented in this work include:
\begin{enumerate}
\item On-line segmentation of the RGB image stream using a state-of-the-art
   semantic segmentation method,
\item fusion of the segmentation onto the 3D mesh recorded by the AR device,
\item presentation of the segmentation results as holograms in the Microsoft HoloLens,
\item and finally demonstration of interaction possibilities with the holograms.
\end{enumerate}

\section{Related Work}

Mixed Reality for HMD is a technology branch that has only gained in relevance in recent years, because technical progress now makes it possible to manufacture devices that meet the hardware requirements and are at the same time light-weight enough to be comfortable. The areas of application are very diverse; however, currently the research interest seems to be concentrated in medicine.
In this discipline the HoloLens is used as a visual extension for physicians, e.g. to represent the position of a tumor \cite{c14}, to project CT scans over the body part to be operated on as help \cite{c15} or to be able to view complex injuries spatially better like a myocardial scar \cite{c7}. The placement of holograms in space is done manually in most applications and takes some time. An extension with automatic placement would be useful and the HoloLens with its spatial mapping capabilities also provides the basis for this.
   
In the work of Evans et al. \cite{c3} a Mixed-Reality-App was designed for the evaluation of the HoloLens, which should guide the user through the assembly instructions. As with a paper manual, there should be the possibility to browse through the individual steps. At each step the required component is indicated and if the component is in the field of view, it is also optically marked. To recognize the different components they first investigated the Spatial Mapping Mesh of the HoloLens. However, it does not provide the resolution to accurately display small objects as mesh and thus allow the application to detect them. They therefore decided to use a marker based tracking.
This workaround is not required in our setting, because the mesh generated by the HoloLens is more accurate and finer since the Windows 10 April 2018 software update (10.0.17134.80) while needing the same computing effort. In addition, our method uses RGB-based semantic segmentation for object detection. Both factors compensate this effect.

Orsini et al. \cite{c12} have presented an application in their work to create a better cooking experience with the HoloLens. When the application is started, the user's environment is searched for food using neural networks. Recipe ideas are proposed based on the detections. Once the user has selected a recipe, instructions are given on how to cook the dish. These instructions are supported by animated holograms. In this way, the user is provided with the most natural assistance possible. Control is carried out via voice commands.
As in our architecture, Orsini et al. use external computation power for semantic perception. In addition, the idea of extending the possibilities of the user through augmented reality corresponds to a main goal in the work of Orsini et al. However, by combining the captured 3D mesh with semantic information, our method is able
to more accurately display object locations to the user.

Chen et al. \cite{c2} have introduced a framework with which they assign semantic properties to the real environment. In order to show the possibilities of mixed reality in connection with these properties, they have programmed a first-person shooter app that shows different bullet holes depending on the material. To detect the semantics of the environment, they use Dense Scene Reconstruction and Deep Image Understanding. Segmentation of the RGB images is performed by a neural network based on CRF-RNN (Conditional Random Fields as Recurrent Neural Network). A Bayesian fusion is used to fuse the semantic labels with the mesh, followed by a refinement using CRFs. The RGB images and the depth images are recorded beforehand with a Kinect V2 sensor, because this provides higher resolution depth images than the HoloLens. All necessary data is preprocessed before starting the application, so that only the information needs to be accessed later. 	
Our approach has many similarities with the work of Chen et al. Also here semantic properties are to be assigned to the real environment and thereby special interactions are to be made possible. However, we use RefineNet to segment 2D images. The advantage of our method is that it works online and does not rely on an external sensor for prior scanning and pre-processing. This makes it much more flexible to use.

\section{Microsoft HoloLens}

The Microsoft HoloLens is powered by a four-core Intel Atom x5-Z8100 processor with 1.04\,GHz, a holographic processing unit (HPU) developed by Microsoft and 2\,GB RAM. Windows 10 is pre-installed as the operating system. A total of 64\,GB memory is available, but a maximum of 900\,MB can be used per app. The built-in battery with 16.5\,Wh is sufficient for 2-3 hours of active use and up to two weeks in standby. This is also the reason for the wait time before the mesh is sent to the server. 
No external localization facilities are required, since the HoloLens features an integrated inside-out tracking. This uses the four ''Environment Understanding Cameras'' and the inertial measurement unit (IMU) to observe prominent points of the environment and to fuse the changes with the IMU's data.

The images are captured as BGRA images. When capturing photos, the API also provides the necessary transformations to locate the capture pose. The transformation and projection matrices are automatically calculated and contain the extrinsic and the intrinsic camera parameters. We capture pictures with a resolution of 896$\times$504\,px.
The Spatial Mapping API allows to change the resolution and time between mesh updates. The mesh has a resolution of 800 triangles per $m^{3}$ and updates every two seconds. Only points that are at least 0.85\.m from the HoloLens are used for updating the mesh to prevent falsification by involving the user's hands. The default scan area used in this application is 4$\times$4$\times$4\,$m^3$ around the initial pose of the HoloLens.

\section{Architectures}

\begin{figure*}
\centering
\includegraphics[width = 0.8\linewidth]{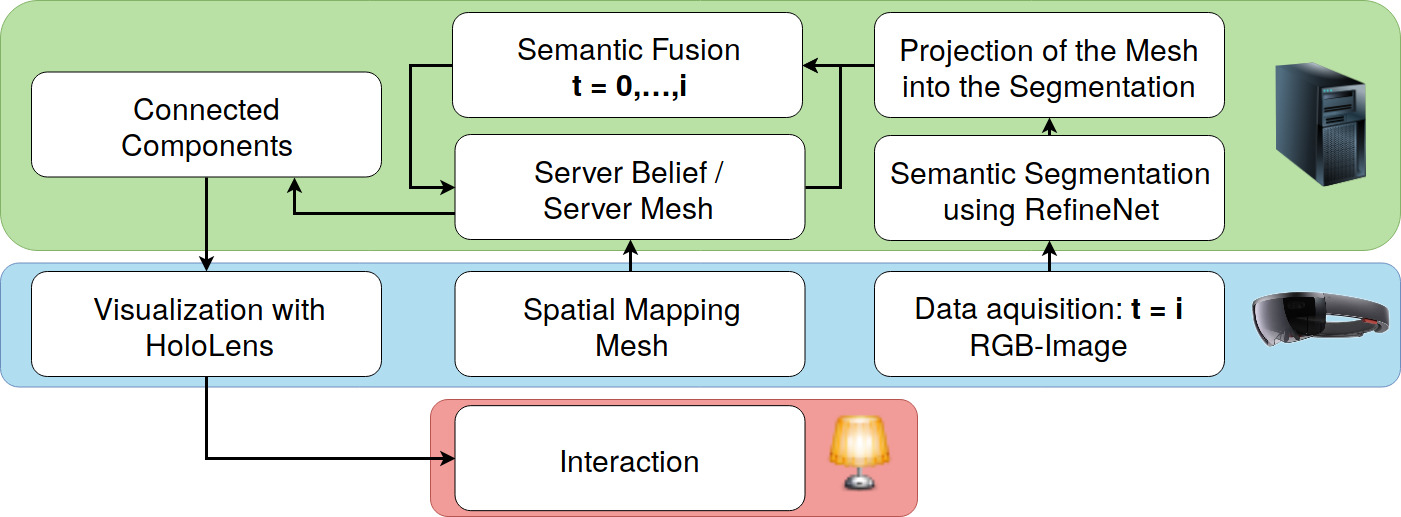}
\caption{Data flow of the proposed architecture. The green part of the diagram is implemented on the server, while blue modules are provided by the HoloLens.}
\label{fig:workflow}
\end{figure*}

The architecture used here is a client-server structure. The Microsoft HoloLens serves as client and a Linux computer as server (see Fig.~\ref{fig:workflow}). Since the on-board processing power of the HoloLens is limited, this
distribution is required.
The app was programmed using the Unity 2017.4.1f1 engine. The server applications were implemented with Python 3.7.1.

Client and server are connected using a WiFi connection. Because the HoloLens was designed as an independent HMD, the wireless connection with lower bandwidth and stability is preferred so as not to restrict the user's freedom of movement compared to a cable connection. For the WiFi connection, TCP sockets are used at both endpoints in order to facilitate communication as easily as possible and to optimize the data stream over the network.

The HoloLens serves to record the required data, to interact with the user and to display the semantic annotations. To ensure these functionalities, the spatial mapping mesh, camera images and transformation matrices are transferred to the server. On the server side, the images are then segmented using the RefineNet method, projected onto the mesh and merged with the previous segmentation. The result is then displayed on the HoloLens to give the user interaction possibilities. More detailed information on the process will follow in the next section. The lamp is controlled by the server using a ROS interface.

\subsection{Data Flow}

In order to provide the user's environment with interaction possibilities by means of segmentation, a constant flow of data from and to HoloLens is required. In this paper, a process as shown in Fig.~\ref{fig:workflow}, is suggested:

When the app is started, it starts to scan the environment and combine the information into a mesh. The resulting Spatial Mapping Mesh is sent to the server after 1000 frames. In our current implementation we fix this mesh as our environment representation and do not update it again. Since the environment is considered static and the mesh is transmitted only once, the user must scan the environment thoroughly before starting the app to provide the application with as accurate a basis as possible.
In the future, this restriction could be lifted by processing geometry updates online as they occur.

Each time step after transmission of the mesh, a new image with its transformation matrices is recorded by the HoloLens. Both are sent to the server. The image is segmented by the server using the pre-trained RefineNet backbone. The mesh is then projected onto the calculated class probabilities per pixel. In order to take into account the results of the previous segmentations, a fusion with all previous time steps takes place.
In this way, the overview of the probability of the classes for each vertex is kept.

In order to generate a segmentation from these probabilities, the maximum of the class probabilities of the vertices must be calculated. As a result, each vertex is assigned a label (''Lamp'', ''Chair'' or ''Unknown'').

Vertices whose class is ''Unknown'' do not interest us in the representation of the segmentation in the HoloLens. After the vertices segmented as ''Unknown'' are filtered out, the rest of the mesh is split into its connected components. This is to separate the individual chairs and lamps from the full mesh. The meshes of the connected components are then filtered by the number of triangles in order to reduce artifacts and false positives in the display and to eliminate meshes that are too small and therefore irrelevant. The number of triangles from which a component is sent varies per class and is determined empirically. By default, a chair must consist of more than 30 and a lamp of more than five triangles.

The connected components thus obtained are sent to the HoloLens for display after a certain number of images (by default five). In addition to the mesh, each component is also annotated with the class to which it belongs in order to transfer the segmentation results to the HoloLens. In the HoloLens, the meshes of the components are created and displayed as independent holograms.
%

When selecting one of the holograms, a ray is cast into the gaze direction and the information about the endpoint of the raycast is saved in world coordinates. The server receives that spatial information and the class name of the hologram, that was hit by the ray. Based on this information it initiates an interaction. The lamp is turned on/off using a microcontroller that controls the lamp's power supply.

\subsection{Rendering}
The Photocapture API of the HoloLens provides two transformation matrices for capturing photos; The \texttt{CameraToWorld} matrix $C$ and the projection matrix $P$. They are calculated from the extrinsic or intrinsic camera parameters.

\begin{figure}
\includegraphics[width = \linewidth]{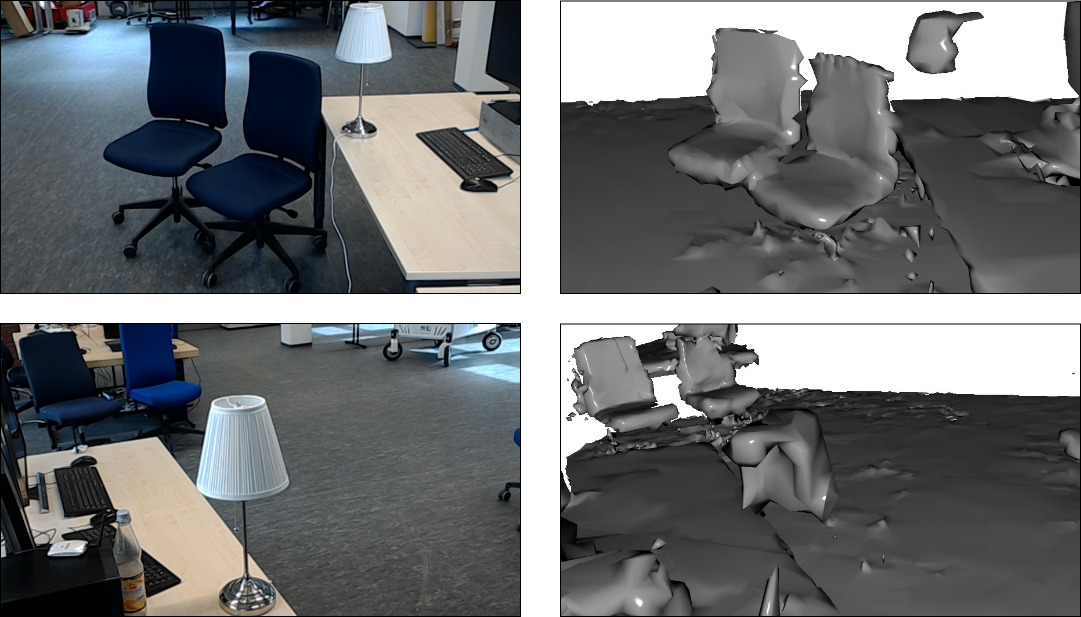}
\caption{Rendering is used to obtain depth maps for a particular viewpoint.
Left: RGB scene as captured by the HoloLens camera. Right: Rendering
of the spatial mapping mesh from the same viewpoint.}
\label{fig:rendering}
\end{figure}

Our rendering module calculates a depth map from the perspective of the camera pose (see Fig.~\ref{fig:rendering}).
Only the class probabilities of the vertices whose Z-distance to the camera has a maximum difference of 0.01\,m to the depth calculated by the render are updated. The remaining vertices are not considered because they are not in the visible area. This ensures that we do not update occluded or invisible vertices. 

\subsection{Semantic Fusion}

The projection of the mesh onto a segmented RGB image gives a probability distribution of the classes for visible vertices. A semantic fusion is necessary in order not to restrict this assignment to the visible area of a single image, but to perform it in the entire mesh. An arbitrary number of segmented images and their projections serve as input. As output a class assignment for each vertex of the mesh should be possible.

Each vertex $v$ in our mesh stores, in addition to its position, a discrete probability distribution $P(L_v = l_i)$ over the set of all classes $l_i \in L$, where $L = \{\text{Lamp},\text{Chair},\text{Unknown}\}$. Each vertex is initialized with an equal distribution over the classes, because there is no \textit{a priori} knowledge.

The output of RefineNet is a per-pixel probability distribution over the class label $P(O_u = l_i | I_k)$, where $I_k$ is the k-th image and $u$ is the pixel coordinate. The visible vertices were determined using the rendering. Now every visible vertex is projected onto an image pixel through the already discussed projection. The probability distribution of these vertices is adjusted with the rule below, which corresponds to a recursive Bayesian filter:%
\begin{equation}
	P(l_i | I_0, ..., I_k) = \eta P(O_{u_{v,k}} = l_i | I_k) P(l_i | I_0, ..., I_{k-1})
\end{equation}%
Here $\eta$ is the normalization factor of the probability distribution.
This methodology follows \cite{c10, c6}.

\section{Experiments}

\subsection{Hardware}
The used Linux server is a desktop PC with an Intel(R) Core(TM) i9-9900K (8$\times$3,6\,GHz), 64\,GB memory and a GeForce\,RTX\,2080. For wireless communication with the HoloLens an ALFA Networks AWUS036NHR USB network adapter with a maximum transfer rate of 150\,MBit is used. The WLAN network in which the HoloLens and the server communicate is hosted by a Windows 10 computer with another ALFA Networks AWUS036NHR. The lamp to be segmented here is the IKEA $\mathring{A}RSTID$ table lamp with a shade diameter of 22\,cm and a shade height of approx. 22\,cm. It is activated using an Arduino microcontroller connected to the Linux server.

\subsection{Semantic Segmentation with RefineNet}

For this work a semantic segmentation of RGB images is required, for which we follow the RefineNet~\cite{c9} method. It is a multi-path refinement network. The idea of the RefineNet pipeline is to merge rough semantic high level features with fine-grained low level features extracted from the intermediate results of the CNNs to create high-resolution semantic feature maps. It outputs a dense score map containing the class probabilities for each pixel. 
We choose two object classes ('Lamp' and 'Chair'), since this paper works as a proof of concept, chairs are one of the objects that appears often in our office environment and a lamp provides an easy to understand and natural interaction possibility. More object classes can be introduced for cluttered environments, since our architecture supports this scalability and RefineNet has shown good segmentation results for cluttered scenarios \cite{c9, c16}.

The parameters for RefineNet were chosen as in the original paper. The learning rate was set to $10^{-5}$ and the updates of the weights were calculated with the Adam optimizer \cite{c8}.
We train our model on the SceneNet~\cite{c11}, a large-scale synthetic dataset
of indoor scenes.
Using a synthetic dataset avoids the need for any human annotation.
It contains more than 5 million synthetic images from more than 15000 trajectories in different synthetic environments. The depth image, the RGB image and an instance segmentation are available for each image. They have a resolution of 320$\times$240 pixels. The trajectories were laid through synthetic spaces populated by objects from ShapeNet. The created scenes are supposed to recreate situations as they might occur in a living room, study or bedroom.  Frames in which chairs or lamps can be found were of interest. In total there are 9985 frames with lamps and 87393 frames with chairs. Using SceneNet gives use the advantage to be able to start training immediately without the need to label a training set ourselves. The network was trained for 20 epochs.

\subsection{Accuracy Study}

Multiple factors affect the accuracy of the system: (1) Inside-out-tracking of the HoloLens, (2) the closeness to reality of the mesh, (3) the 3D to 2D semantic fusion and (4) the trained neural network. Through this interweaving of parts it is difficult to evaluate each part. We therefore evaluated those parts as a whole. 
36 frames were taken from multiple application runs as RGB images and labeled manually to obtain ground truth data for the segmentation. Based on these we calculate four common metrics for semantic segmentation and scene parsing evaluations to examine the variations of pixel accuracy and region intersection over union (IU)~\cite{c18}. Let $n_{ij}$ be the number of pixels of class $i$ predicted to belong to class $j$, where there are $n_{cl}$ different classes, and let $t_i=\sum_{j}n_{ij}$ be the total number of pixels of class $i$. We then obtain the following metrics:

\begin{enumerate}
\item Pixel accuracy: $\sum_{i}n_{ii}/\sum_{i}t_{i}$,
\item mean accuracy: $(1/n_{cl})\sum_{i}n_{ii}/t_{i}$,
\item mean IU: $(1/n_{cl})\sum_{i}n_{ii}/(t_{i}+\sum_{j}n_{ji}-n_{ii})$, and
\item frequency weighted IU:\\ $(\sum_{k}t_{k})^{-1}\sum_{i}t_in_{ii}/(t_{i}+\sum_{j}n_{ji}-n_{ii})$.
\end{enumerate}

\renewcommand{\arraystretch}{1.2} 
\newcolumntype{C}[1]{>{\centering\let\newline\\\arraybackslash\hspace{0pt}}m{#1}}
\begin{table}
\caption{Quantitative results on indoor scenes dataset}
\label{tab:results}
\centering
 \begin{tabular}{C{0.22\linewidth}|C{0.135\linewidth}|C{0.135\linewidth}|C{0.13\linewidth}|C{0.13\linewidth}} 
    & pixel acc. & mean acc. & mean IU & f.w. IU \\\hline
  RefineNet w. Semantic Fusion & 88.05 & 76.86 & 65.76 & 80.93 \\
\end{tabular}
\end{table}

\begin{table}
\caption{Measurement data for runtime analysis. An image of the used resolution (896$\times$504) needs around one second to send.}
\label{tab:results_time}
\centering\begin{tabular}{C{0.28\linewidth}||C{0.28\linewidth}|C{0.28\linewidth}} 
Waiting time for photo [frames] & network load [MB/s] & max. backlog [\# images]\vspace{0.1cm}\\ \hline
30		&2,15	&50\\ 
60		&1,80	&3\\
100	&1,09	&0
\end{tabular}
\end{table}

The values given in Tab.~\ref{tab:results} are within an acceptable range and together with the qualitative results in the next section we can state that the segmentation works sufficiently well for the given task. Projection onto a coarse mesh therefore does not significantly degrade the quality of our result.

\begin{figure}
\centering
\includegraphics[width = \linewidth]{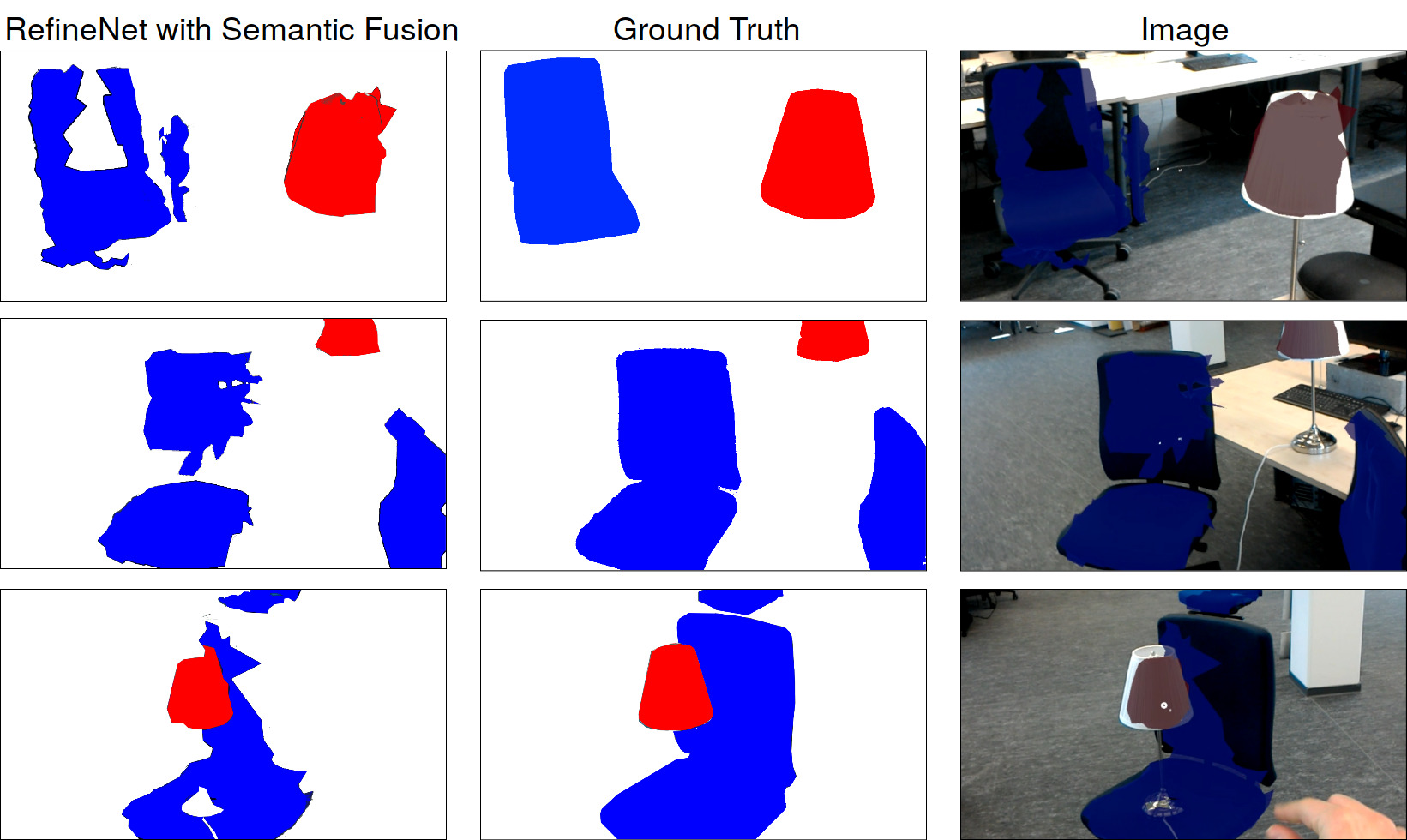}
\caption{Examples from the semantic segmentation for each column from the left to right: Segmentation with RefineNet and Semantic Fusion on 3D, Ground Truth, Input image}
\end{figure}

\subsection{Qualitative Results}

\begin{figure}[t]
\centering
\includegraphics[width = \linewidth]{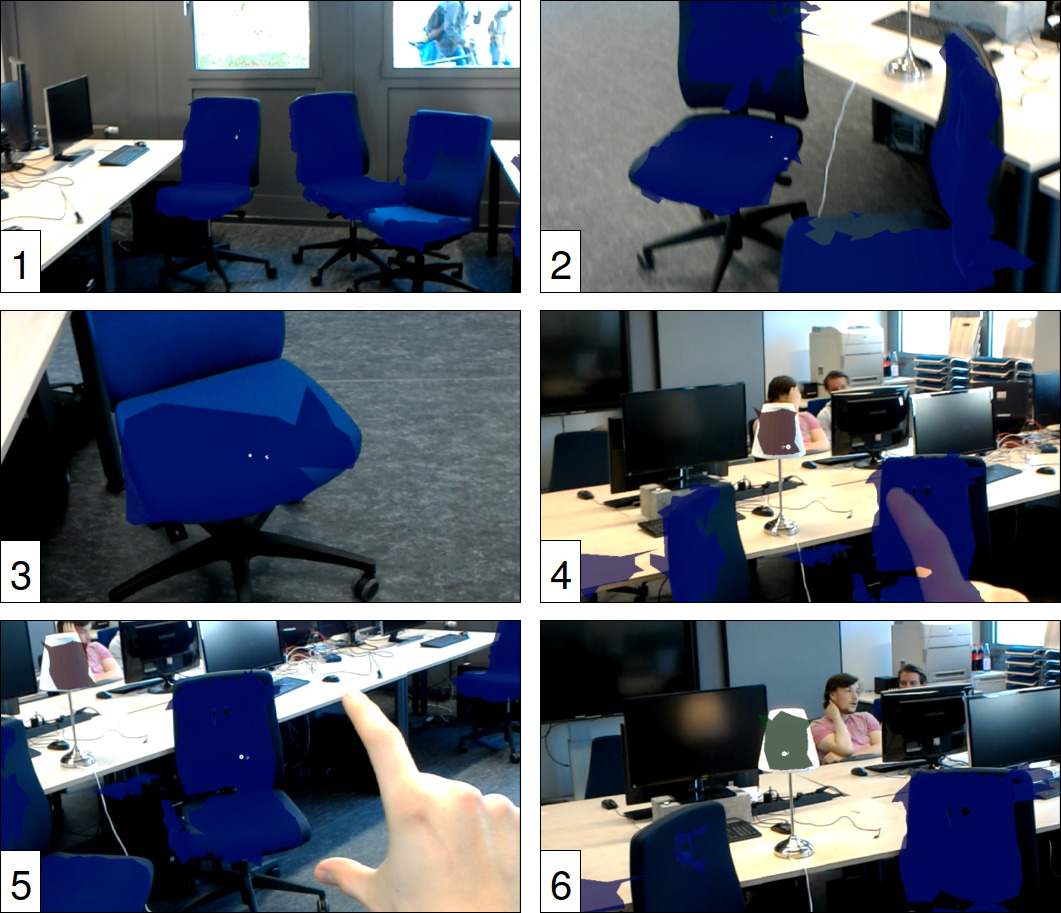}
\caption{Full mapping run. Six example frames are shown extracted from a video
 available at \url{https://www.ais.uni-bonn.de/videos/ECMR_2019_Schuett/schuett_ecmr.mp4}.}
\label{fig:6Frames}
\end{figure}

In Fig.~\ref{fig:6Frames} several frames have been taken from a run of the app
available as video\footnote{\url{https://www.ais.uni-bonn.de/videos/ECMR_2019_Schuett/schuett_ecmr.mp4}}.
The video was taken after a first tour with a scan of the environment. For the first 1000\,frames of each application run the wireframe of the mesh is still displayed to signal that the environment is being captured (see Fig.~\ref{fig:4Frames_start} top left). The following provides a detailed explanation of the frames:

\begin{itemize}
\item[1:] The lamp placed on the chair was detected and displayed by segmentation. The chair is not perfectly captured because it has not yet been viewed from this perspective and the lamp has partially obscured the chair.
\item[2-4:] The surrounding office chairs were well segmented and fused to the mesh. One can see an misalignment of the position between hologram and reality. This results from the imperfect localization of the HoloLens and becomes larger.
\item[5:] The lamp is targeted with the cursor (white dot) and selected using the air tap gesture.
\item[6:] The lamp has turned on. Due to the selection the mesh of the lamp has changed from red (not selected) to green (selected).
\end{itemize}

While precise enough for semantic fusion, the low resolution of the mesh becomes
especially apparent during the visualization of the results.
Because of its relatively small size and special shape, the lamp is usually only partially scanned and does not have the conical shape as in reality. In the case of chairs, this problem is not so important because their size allows a realistic representation in the mesh. However, a good and distinct representation of the interaction object must be given priority.
With the current mesh resolution provided by inside-out tracking devices, we
recommend to show only abstract icons as interaction cues instead of full meshes in order to visualize the semantics. The possibility of interaction would be retained and it would be more aesthetically pleasing for the user.

\begin{figure}[t]
\centering
\includegraphics[width = \linewidth]{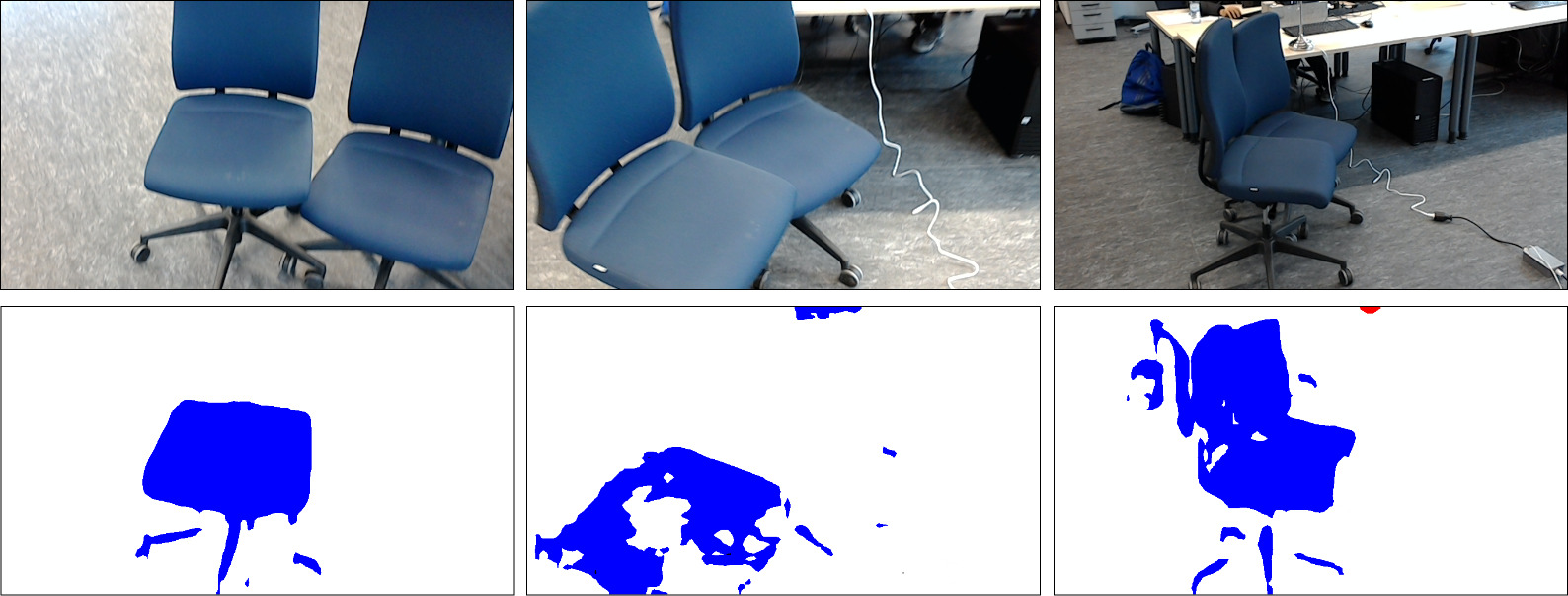}
\caption{Problematic cases for segmentation. Top: RGB input, bottom: Segmentation results.
In the pictures on the left and in the middle you can see that our segmentation fails for partial views of office chairs, but gives very good results as soon as the whole chair is visible again (right).}
\label{fig:probleme_segmentierung}
\end{figure}

In most app runs it could be seen that the correct segmentation of the chairs across many frames was a big problem. Tests carried out as a result have shown that the trained RefineNet has problems with correctly segmenting all areas of an office chair when they are only partially visible. This is shown in Fig.~\ref{fig:probleme_segmentierung} and seen in the lower pictures in Fig.~\ref{fig:4Frames_start}. Due to this problem, the results of the semantic fusion for chairs are very bad if many pictures were taken in the near vicinity of these chairs. The lamp does not show these bad segmentation results, because its small size and positioning allows it to always capture them completely in the image in most cases.

The idea of our solution approach is not to update the results of segmentation, which have ''Unknown'' as the most likely class and are less than two meters from the camera position. In this way, the semantic fusion result is not affected by false negatives. Problematic is a resulting high confidence of the classes ''lamp'' and ''chair'' and increased false positives by a missing update of the confidence of these vertices. It is therefore recommended to ensure that objects are viewed from a distance of more than two meters in order to update and adjust the false positive classified vertices.

\section{Conclusions}
In this work, we demonstrated scanning of the environment using the HoloLens
and creating a semantic mesh representation through the usage of segmentation
and probabilistic fusion.
Our method runs on-line with one frame per second (Tab.~\ref{tab:results_time}). The resulting semantic
interaction cues can be used for a variety of applications, here demonstrated
by controlling a lamp.
The fusion was implemented using a method corresponding to a recursive Bayesian filter, the presentation of the results using connected components and the interaction through the selection gesture.
Our approach could be used for human robot interaction/cooperation, because the ability to show the robot the interaction object in reality gives the user a more intuitive way to control it.

The HoloLens has weaknesses that limit its usage in the robotics field. The Spatial Mapping Mesh has a resolution that is too low to allow a large amount of detail. The area where holograms are visible is very small with 34$^{\circ}\times$19.5$^{\circ}$ and does by far not cover the user's field of view, so that the immersive feeling is often lost.

In the future, some of the assumptions made should be generalized to evaluate the impact on segmentation results. Examples of this are the assumption of a static world and the use of pre-processed mesh data as information. The use of the raw data to create the mesh by ourselves on the server could be an opportunity to make the server-sided mesh much more detailed and thus to display the segmentation results more accurately. In addition, the semantic segmentation
method should be improved, for example by creating synthetic training data that
more closely fits the situation at hand. 


\bibliographystyle{IEEEtran}
\bibliography{IEEEabrv,root}

\end{document}